\documentclass[sigconf]{acmart}
\usepackage{multirow}
\usepackage{geometry}
\usepackage{graphicx}
\usepackage{enumitem}
\usepackage{subfig}
\usepackage{caption}
\AtBeginDocument{%
  \providecommand\BibTeX{{%
    \normalfont B\kern-0.5em{\scshape i\kern-0.25em b}\kern-0.8em\TeX}}}

\copyrightyear{2021} 
\acmYear{2021} 
\setcopyright{acmcopyright}\acmConference[CIKM '21]{Proceedings of the 30th ACM International Conference on Information and Knowledge Management}{November 1--5, 2021}{Virtual Event, QLD, Australia}
\acmBooktitle{Proceedings of the 30th ACM International Conference on Information and Knowledge Management (CIKM '21), November 1--5, 2021, Virtual Event, QLD, Australia}
\acmPrice{15.00}
\acmDOI{10.1145/3459637.3482424}
\acmISBN{978-1-4503-8446-9/21/11}

\newcommand{\ie}{\emph{i.e., }}
\newcommand{\eg}{\emph{e.g., }}

\newcommand{\aka}{\emph{aka. }}

\begin{document}
\fancyhead{}
\title{DisenKGAT: Knowledge Graph Embedding with Disentangled Graph Attention Network}
\author{Junkang Wu$^{1\dagger}$, Wentao Shi$^{1}$, Xuezhi Cao$^2$, Jiawei Chen$^{1*}$ \\Wenqiang Lei$^3$, Fuzheng Zhang$^2$, Wei Wu$^2$, Xiangnan He$^1$}
\affiliation{\institution{$^1$University of Science and Technology of China, $^2$Meituan, 
$^3$National University of Singapore.}
\country{}} 
\email{{jkwu0909@mail., shiwnetao123@mail., cjwustc@, hexn@}ustc.edu.cn}
\email{{caoxuezhi@, zhangfuzheng@, wuwei30@}meituan.com, wenqianglei@gmail.com}


\thanks{$*$ Corresponding author.\\
$\dagger$ Work done at Meituan.} 

\renewcommand{\authors}{Junkang Wu, Wentao Shi, Xuezhi Cao, Jiawei Chen, Wenqiang Lei,
Fuzheng Zhang, Wei Wu, Xiangnan He}

\begin{abstract}
Knowledge graph completion (KGC) has become a focus of attention across deep learning community owing to its excellent contribution to numerous downstream tasks. Although recently have witnessed a surge of work on KGC, they are still insufficient to accurately capture complex relations, since they adopt the single and static representations. In this work, we propose a novel Disentangled Knowledge Graph Attention Network (DisenKGAT) for KGC, which leverages both \textit{micro-disentanglement} and \textit{macro-disentanglement} to exploit representations behind Knowledge graphs (KGs). To achieve micro-disentanglement, we put forward a novel relation-aware aggregation to learn diverse component representation. For macro-disentanglement, we leverage mutual information  as a regularization to enhance independence. With the assistance of disentanglement, our model is able to generate adaptive representations in terms of the given scenario. Besides, our work has strong robustness and flexibility to adapt to various score functions. Extensive experiments on public benchmark datasets have been conducted to validate the superiority of DisenKGAT over existing methods in terms of both accuracy and explainability.The code is available at \url{https://github.com/Wjk666/DisenKGAT}. 
\end{abstract}



\maketitle
\begin{figure}[!t]
    \centering 
    \includegraphics[width=0.4 \textwidth]{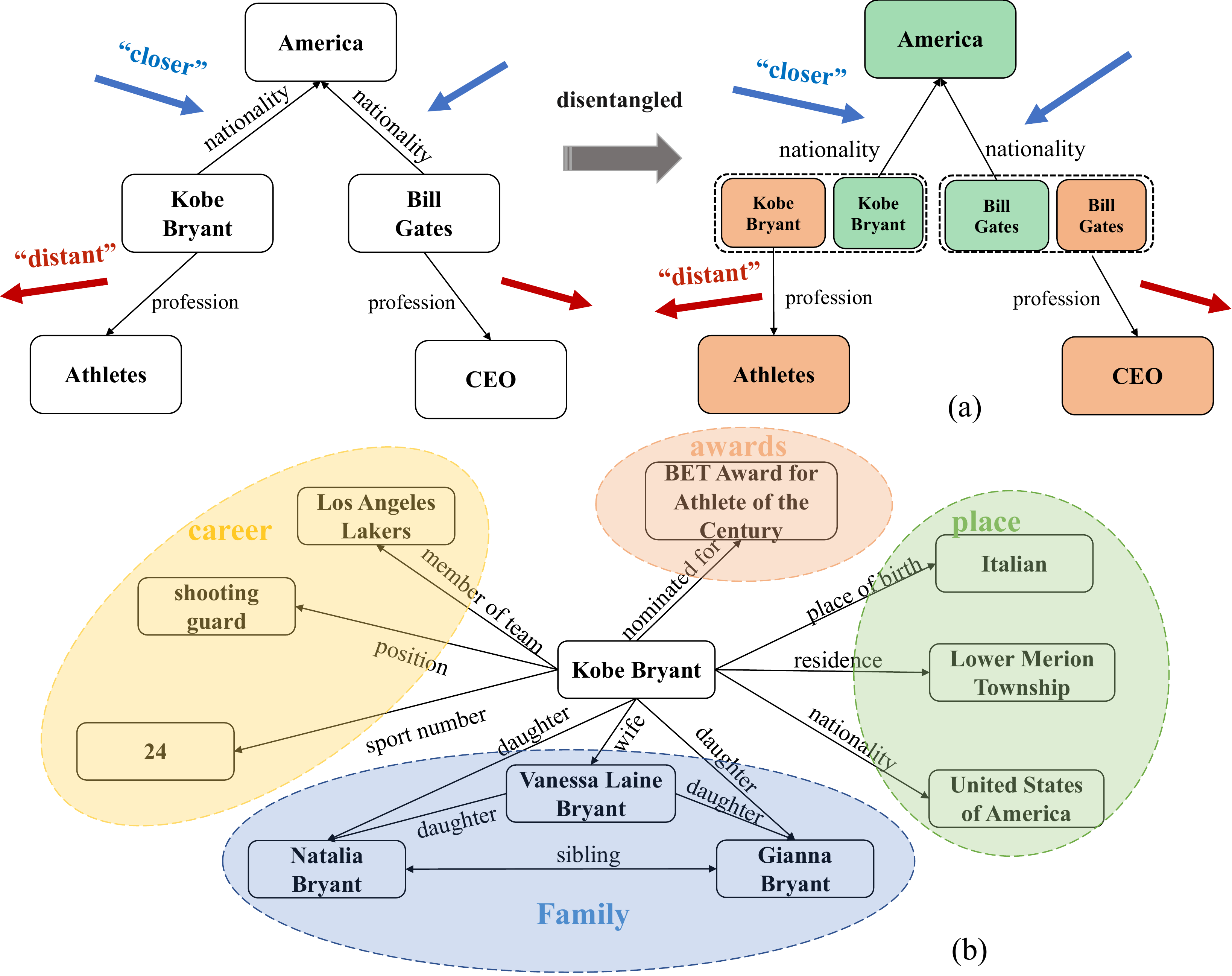} 
    \caption{ An example of the "Kobe Bryant" and "Bill Gates" neighboring distribution, where different color depicts a  "topic" or "cluster". It is clear to figure that an entity always plays multiple roles in different scenarios.} 
    \label{fig:kobe} 
\end{figure}
\begin{figure*}[!ht] 
    \centering 
    \includegraphics[scale=0.33]{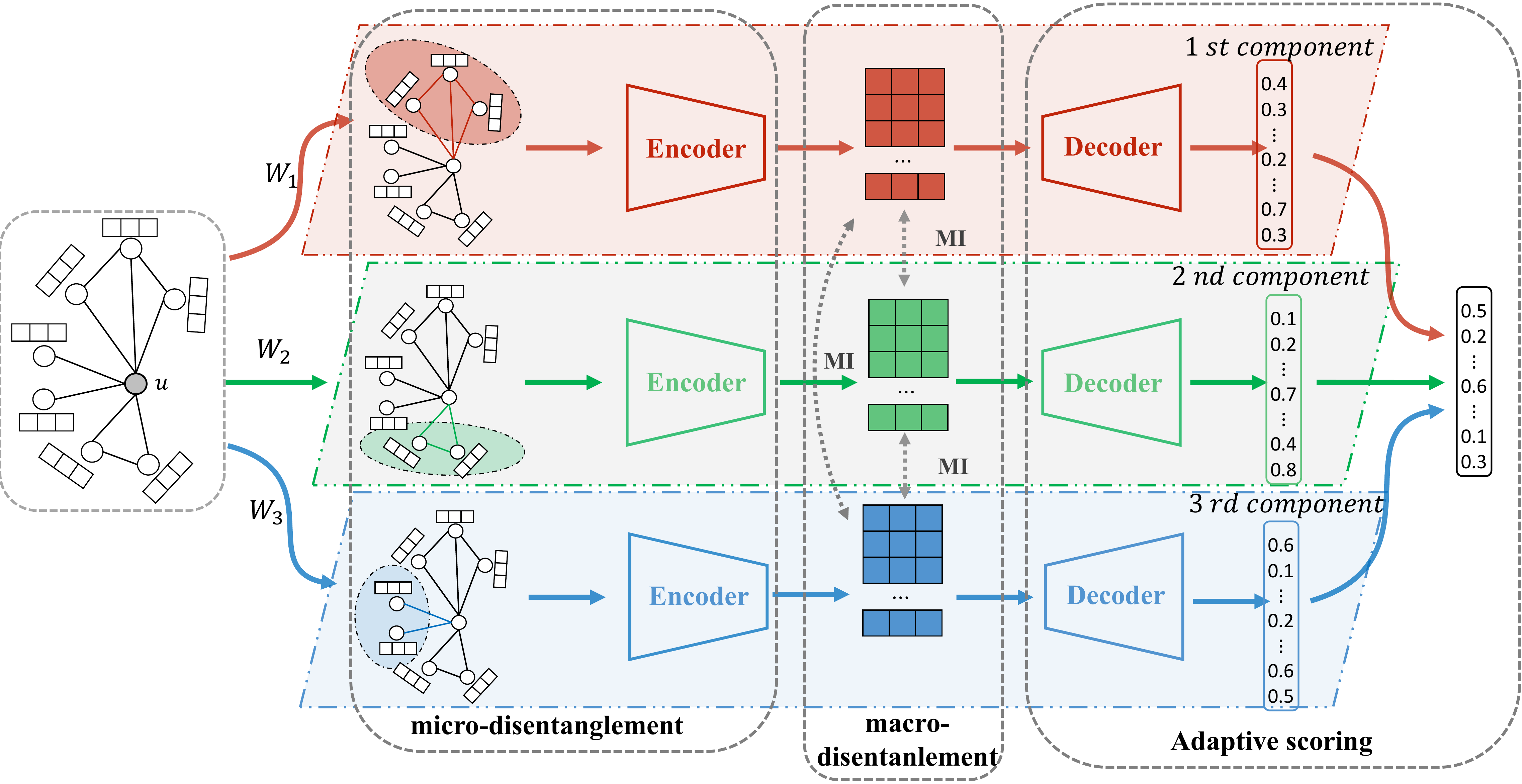} 
    \caption{An architecture overview of our DisenKGAT. The whole model contains three key modules: (1) relation-aware aggregation, (2) independence constraint, and (3) adaptive scoring. Each channel (color) represents a specific component and different components keep individuals with each other by the correlation restriction. The final prediction result depends on all components of the entities and is adaptive to the given scenario (relation). Best viewed in color.} 
    \label{fig::model} 
\end{figure*}
\section{Introduction}
In recent years, knowledge graphs (KG) aroused considerable research interest since they changed the paradigm for numerous state-of-the-art natural language processing solutions and enhanced the AI-related applications. Representative applications include \eg dialogue generation \cite{he2017learning}, question answering \cite{madotto2018mem2seq}, recommender systems \cite{zhang2016collaborative}, etc. Although modern KGs such as FreeBase \cite{bollacker2008freebase}, WordNet \cite{miller1995wordnet}, Nell \cite{mitchell2018never} have already contained millions of entities and triplets, they are still far away from complete due to the constantly emerging new knowledge. Therefore, knowledge graph completion (KGC), which predicts whether a new given triple is valid or not, still attracts much research attention. 

In literature, most state-of-the-art KGC models are based on knowledge graph embedding, which focuses on learning low-dime-\\nsional embedding for entities and relations. These models could be classified into three types, translational models \cite{bordes2013translating, wang2014knowledge, lin2015learning}, bilinear transformation models \cite{yang2014embedding, nickel2016holographic, trouillon2016complex} and convolutional network based models \cite{dettmers2018convolutional, jiang2019adaptive, nguyen2017novel}. Although these models have achieved excellent performance, they still ignored the meaningful graph context such as vast structural information. In this way, Graph neural network based models \cite{schlichtkrull2018modeling, marcheggiani2017encoding, shang2019end, vashishth2019composition} become an active research focus for knowledge graph completion. Inspired by the various graph aggregation mechanisms \cite{kipf2017semi, velickovic2018graph}, they make use of layer-wise propagation to collect neighboring entity embedding incorporating the relation representation. In addition, CompGCN \cite{vashishth2019composition} proposes a novel framework for combining the aggregation ability of the GCN and the representative score function to construct the encoder-decoder paradigm in knowledge graph completion.

Although significant progress has been made by the combination of the graph structural information and semantic prediction, static representation still limits the flexibility and expressiveness of the embedding in particular in complex one-to-many (1-to-N), many-to-one (N-to-1), and many-to-many (N-to-N) relations. Typically, as shown in Figure \ref{fig:kobe} (a), "nationality" is a many-to-one relation --- everyone has a unique nationality and numerous people have an identical nationality. Take the query \textit{(Kobe, profession, ?)} and query \textit{(Bill Gates, profession, ?)} as an example, owning to the common nationalities (``America'') of ``Bill Gates'' and ``Kobe'', existing methods will naturally pull the representations of \textit{Kobe} and \textit{Bill Gates} closer to capture such identity. However, in terms of the relation "profession", where the two people indeed exhibit quite a difference, such static and similar representations will obviously affect the link prediction performance. In other words, existing methods can not generate a different representation while the scenario changes dynamically.

According to the issue mentioned above, we summarize that (i) simply aggregating neighborhood entities fails to effectively model the critical relationship that is significant in characterizing the local information in a specific scenario. We argue that entities and relations when engaged in different graph contexts, might exhibit extremely distinct meanings. (ii) They ignore the entanglement of the latent factors behind the entities embeddings while an entity may have multiple aspects and various relations focus on distinct aspects of entities. As the example in Figure \ref{fig:kobe} (b), the entity "Kobe Brant" is associated with different task components such as "Family", "Career", "Place" and "awards". Assume that in the prediction task of "Kobe's son", it's supposed to pay more attention to the component family such as wife and daughters rather than the irrelevant component about career or awards.  (iii) In the light of the above two points, these methods result in low interpretability and non-robustness, i.e., they are prone to overreact to an irrelevant neighbor which does not fit the current scenario and wipe out the rich structural and context information for KGs itself.

In this work, we provide a novel Disentangled Knowledge Graph Attention network (DisenKGAT) that is able to learn the disentangled representation of the entities in terms of the property of the knowledge graph and has strong robustness in several kinds of representative score functions. To tackle the issue of entanglement of latent factors, we propose a mechanism that incorporates the micro-and macro-disentanglement. More specifically, As for the micro-disentanglement, that is, each entity contains multiple diverse components, our model leverages a dynamic assignment mechanism to collect neighboring entities' representation similar to the current scenario. Furthermore, as for the consideration of macro-disentanglement, that is, different components should separate clearly, we utilize mutual information regularization to ensure each component be independent of each other. Besides, the learned disentangled representation can be along with different kinds of score functions which is also proof of strong robustness.

we summarize the main contributions of our work as follows:
\begin{itemize}
    \item We propose a novel Disentangled knowledge graph convolutional network to learn the disentangled embedding for knowledge graphs. And the proposed encoder model can be potentially generalized to other score functions in the task of knowledge graph completion. To the best of our knowledge, this is the first attempt to leverage the disentangled representation in knowledge graph completion.
    \item To achieve micro-disentanglement of entities, we propose a special relation-aware aggregation mechanism, which aggregating the neighborhood information at a semantic level while ensuring the consistency of the adaptive scoring part of the model.
    \item To achieve macro-disentanglement of the component distribution level, we propose a mutual information-based regularization to reduce intra-component correlation.
    \item We conduct extensive experiments on real-world benchmark datasets to validate the effectiveness of our proposal. Our DisenKGAT not only outperforms state-of-the-arts but also has strong robustness and interpretability.

\end{itemize}

\section{Related work}
\subsection{Knowledge Graph Embedding}
The fundamental point of knowledge graph representation learning is to learn how low-dimensional distributed embedding of entities and relations. These models could be roughly divided into three categories: Distance-based model \cite{bordes2013translating, wang2014knowledge, lin2015learning} , semantic models \cite{yang2014embedding, nickel2016holographic, trouillon2016complex}, and neural network based models \cite{dettmers2018convolutional, jiang2019adaptive, nguyen2017novel}. Most of these KG embedding methods build a paradigm that defining a score function at first to measure the plausibility of the given triplets and then making sure positive samples have higher scores than negative samples. Despite the success of these approaches, there still exists lacking consideration for knowledge graph structural information. Meanwhile, graph neural network based KG \cite{schlichtkrull2018modeling, marcheggiani2017encoding, shang2019end, vashishth2019composition} models try to use more graph structure information and gradually develop it into a mature and more flexible aggregation pattern adapted to the characteristics of the complex knowledge graph. However, they are still learning a static representation for each entity and relation which can not deal with the 1-to-N, N-to-1, and N-to-N relations.

Motivated by the aforementioned observations, the problem has received scant attention in the research literature, but from distinct views. \cite{wang2014knowledge, lin2015learning, ji2015knowledge} provide the relation-specific projections to project an entity into a different latent space related to the given relation. Coke \cite{wang2019coke} utilizes the transformer to exploit the contextualized and dynamic representations for entities and relations. FAAN \cite{sheng2020adaptive} proposes an adaptive attention network and a transformer encoder which could help to adaptively match references with queries. It seems that the mainstream method of tackling the issue of static representation borrows the capability of Transformer in capturing the dynamic KG embeddings. It has even shown that the aggregation mechanism in Transformers \cite{Vaswani2017transformer} is similar to dealing with a complete graph by GNNs which could benefit entity representation learning and graph structural mining collaboratively. Although similar phenomena have been touched upon in previous work, there is a little formal discussion about knowledge graph embedding with disentangled representation to investigate latent factors behind the entities.

\subsection{Disentangled Representation Learning}

 Disentangled representation learning, which aims to learn embedding including various separate components behind the data, has get significant progress in numerous fields, such as texts \cite{john-etal-2019-disentangled}, images \cite{NIPS2016_7c9d0b1f} user behaviors \cite{NEURIPS2019_a2186aa7} and recommendation \cite{DGCF19}. In the field of the graph, DisenGCN \cite{ma2019disentangled} is the first work that tends to achieve the micro-disentanglement at the sample level but lacks the consideration of macro-separability. IPGDN \cite{liu2020independence} and ADGCN \cite{zheng2021adversarial}, which utilize the Hilbert-Schmidt Independence Criterion (HSIC) and adversarial learning respectively, further extend to achieve micro-and macro- disentanglement to ensure independence properties. And \cite{zheng2021adversarial} is the pioneer work which put forward the concept of micro- and macro- disentanglement explicitly. The above models develop the disentanglement idea in homogeneous networks, while DisenHAN \cite{wang2020disenhan} focus on Heterogeneous networks. Despite the success of disentangled learning attempts in graphs, how to apply it in complex knowledge graph still remain challenging. Here, our work focus on learning disentangled representation in KG, which generates adaptive embedding in the light of the scenario.

\section{PRELIMINARY}
In this section, we first present some important concepts related to knowledge graph and mutual information. Then we give the formal definition of our task.

\textbf{Knowledge Graph.} Knowledge graphs (KGs) is defined as a directed graph that stores structured information of real-world entities and facts. Let $\mathcal{G}=\{\mathcal{V}, \mathcal{R}, \mathcal{L}\}$ be an instance of a KG, where $\mathcal{V}$, $\mathcal{R}$ and $\mathcal{L}$ denote the entity (node), relation and edge (fact) sets respectively. Each edge $e \in \mathcal{L}$ is presented as a triple $(h,r,t)\in \mathcal{V}\times \mathcal{R} \times \mathcal{V}$, describing there is a relationship $r \in \mathcal{R}$ from head entity $h$ to tail entity $t$. For example, the triple \textit{(Kobe Byrant, daughter, Natalia Bryant)} describes that  \textit{Kobe Byrant} has a daughter named \textit{Natalia Byrant}.


\textbf{Mutual information.} Mutual information (MI) is a fundamental quantity for measuring the dependence of two random variables. The mutual information between the random variables $X$ and $Z$ is defined as follow:
\begin{equation}
    \mathcal{I}(x,z)=\mathbb{E}_{(p(x,z))}[log\frac{p(x,z)}{p(x)p(z)}]
\end{equation}
where $p(x,z)$ is the joint probability distribution of $X$ and $Z$, while $p(x)$ and $p(z)$ are the corresponding marginal distribution. In contrast to correlation, mutual information is able to capture non-linear dependencies between variables and thus acts as a standard measure of true dependence \cite{belghazi2018mutual}.

\textbf{Task Formulation.} The task of knowledge graph completion involving inferring the missing edges in the knowledge graph $\mathcal{G}$, \ie predicting the target entity of a given query of head entity and relation $(h,r,?)$ \footnote{Some of the related work formulate the problem as inferring the missing head entity given a tail entity and a a relation $(?,r,t)$. In fact, the two problems are reducible to the same problem --- learning a proper score function $\psi(h,r,t)$.}. Specifically, the task is usually formulated as a ranking problem, which aims at learning a score function  $\psi(h,r,t):\mathcal{V}\times\mathcal{R}\times\mathcal{V}\to\mathbb{R}$ that assigns a higher score for valid triplets than the invalid ones.

\section{Disentangled Knowledge Graph Attention Network}
In this section, we present our disentangled knowledge graph attention network (DisenKGAT), an end-to-end deep model that is able to learn disentangled entities representation. Overall, the architecture of DisenKGAT is illustrated in Figure \ref{fig::model}, consisting of four modules: (1) Disentangled Transformation (Section \ref{se:tr}): we first project original features of each entity into $K$ different latent spaces, which encode distinct semantic of the entity; (2) Relation-aware Aggregation (Section \ref{se:ag}): we leverage relation-aware attention mechanism to identify relevant neighbors for each factor and then correspondingly guide the neighbor information aggregation; (3) Independence constraint (Section \ref{se:ic}): mutual information regularization is introduced to enhance the independence between each disentangled components; (4) Adaptive Scoring (Section \ref{se:as}): we generate the final prediction  by adaptively combining the results from each component according to the given scenario (\ie relation).

\subsection{Disentangled Transformation}
\label{se:tr}
Our model aims to learn disentangled representation where each component contains distinct semantic. Specifically, for each entity $u$, we would like its representation to be composed of $K$ independent components, \ie, $e_u=[h_{u,1},h_{u,2},...,h_{u,K}]$, where $h_{u,k} \in \mathbb{R}^{\frac{d_{embed}}{K}}$ depicting the $k$-th aspect of the entity $u$. To achieve this goal, we first project the feature vector $x_u$ into different latent spaces in order to make each component extract distinct semantic from the node feature:
\begin{equation}
    h^0_{u,k}=\sigma(W_k\cdot x_u)
\end{equation}
where the initial embeddings $\{h^0_{u,k}\}^K_{k=1}$ are obtained from $x_i$ through K distinct projection matrix $W=\{W_1, W_2,\cdots, W_K\}$, $\sigma$ is the activation function and $x_u$ is the feature of the entity $u$. Note that here we do not use any normalization such as BatchNorm \cite{ioffe2015batch} or LayerNorm \cite{ba2016layer}, as we empirically find it ineffective in our task.


\subsection{Relation-aware Aggregation for Micro-disentanglement}
\label{se:ag}
The feature vector is typically incomplete in real-world and is insufficient to generate a satisfactory embedding. As such, to comprehensively characterize each component of entities, neighborhood aggregation is required so that the learning of each entity can benefit from the rich information from the neighborhoods. However, in a disentangled knowledge graph learning task, we should not use all the neighbors when re-constructing node component $h_{u,K}$ as only a subset of neighbors carry the useful information for the aspect $k$. Take the entity ``Kobe Bryant" (as shown in Figure \ref{fig:kobe}) as an example, when we would like to learn the component of "Family", we must highlight those neighbors with relation ``daughters" and ``wife'' while down weight the contribution of others. In order to more accurately describe the $k$-th aspect of an entity, we must answer the following important question: How to identify the subset of relevant neighbors that are actually connected by the entity $u$ due to the fact $k$? Although recent years have seen a neighborhood routing strategy for homogeneous disentangled graph learning \cite{ma2019disentangled}, it is no longer suitable for knowledge graph as it would neglect crucial information --- the edge relation, which provides explicit evidence of why two entities are connected and naturally plays a key role in aggregation.


For this purpose, in this paper, we propose a novel relation-aware aggregation strategy. We first conduct component-level interaction and then leverage relation-aware attention mechanism to adaptively combine useful information from relevant neighbors for each component. For each entity $u$ and each component $k$, we conduct the following interaction to generate message from the specific neighbor $v$ with the relation $r$:


\begin{equation}
\label{mess_agg}
    m_{(v,k,r)} = \phi(h_{v,k}, h_r,\theta_r)
\end{equation}
\begin{equation}
\label{formulation:theta}
    \theta_r=W_r=diag(w_r)
\end{equation}

Here, $\phi:\mathbb{R}^d \times \mathbb{R}^d \times \mathbb{R}^d \to \mathbb{R}^d$ is a composition operator; $m_{(v,r)}$ denotes the message aggregated from neighbor $v$ with relation $r$; and $\theta_r \in \mathbb{R}^d$  denotes the relation-specific projection matrix. Borrowed the idea from \cite{yang2014embedding} , here $\theta_r$ is restricted to be a diagonal matrix for efficiency. Note that the implementation of $\phi$ can refer to the recent work on vanilla knowledge graph embedding. In this paper, we implement the composition operator $\phi$ in a variant way, with four kinds of approaches including subtraction, multiplication, cross interaction and circular-correlation (detailed formula refer to Section \ref{sec:vis}). This result validates the generalization of our method, which can enjoy the merits of existing methods.

In order to better capture the relevance between $u$ and $v$ in each component space, we leverage a relation-aware attention mechanism to infer the importance of each neighbor in aggregation. Here we simply adopt the dot-product(DP) attention based on the hypothesis that the more similar the entity $u$ and the neighbor $v$ are in the $k$-th component in terms of their relation $r$, the more likely the factor $k$ is to be the reason for the connection.
\begin{equation}
    \begin{aligned}
    \alpha_{(u,v,r)}^k &=softmax((e_{u,r}^k)^T\cdot e_{v,r}^k) \\
    &=\frac{exp((e_{u,r}^k)^T\cdot e_{v,r}^k)}{\sum_{(v^{\prime},r)\in \hat{N}(u)}exp((e_{(u,r)}^k)^T\cdot e_{(v^{\prime},r)}^k)}
    \end{aligned}
\end{equation}

where $e_{u,r}^k=c_{u,k}\circ \theta_r$, represents the $k$-th relation-aware component representation of entity $u$ in a specific relation-aware subspace. $\theta_r$ is shared with the $\theta$ in Formula (\ref{formulation:theta}) in order to ensure the consistency of the semantic subspace. $\hat{N}(u)$ denotes the neighboring entities $N(u)$ including $u$ itself.

Obtained the attention score, we could aggregate the representation from neighbors in each component respectively. Let $h_{u,k}^{l+1}$ denotes the $k$-th component representation of the entity $u$ obtained after $l$ layers which is defined as:
\begin{equation}
   h_{u,k}^{l+1}=\sigma(\sum_{(v,r)\in \hat{N}(u)}\alpha_{(u,v,r)}^k \phi{(h_{v,k}^l,h_r^l, \theta_r)})
\end{equation}

Similarly, here $h_r^l$ denotes the representation of a relation $r$  in the $k$-th layer. We also update $h_r^l$ with layer-wise linear transformation with parameters $W_{rel}^l$ as:
\begin{equation}
    h_r^{l+1}=W_{rel}^l\cdot h_r^l
\end{equation}

In summary, the relation-aware aggregation strategy guarantees that $h_{u,k}^{l+1}$ is prone to be aggregated into a tight area in each component subspace. \textbf{We refer to \cite{zheng2021adversarial}} and would like to name such disentangled mechanism as \textit{micro-disentanglement}.

\subsection{Independence Constraint for Macro-disentanglement}
\label{se:ic}
In the complex multi-relational graph such as knowledge graph, we expect that there should be a weak dependence among the different component representations. Similar objectives have been proposed in different views such as distance correlation and Hilbert-Schmidt Independence Criterion (HSIC) \cite{liu2020independence}, but we argue that controlling the linear correlation is insufficient, as the nonlinear dependence commonly exists especially in complex knowledge graphs. Mutual information, which is a fundamental quantity for measuring non-linear dependence of two random variables, sheds a light on achieving comprehensive disentanglement, \ie directly controlling the mutual information between the disentangled components would be more a promising solution for disentangled graph representation learning.

Although the idea is straightforward, the solution is challenging. Directly optimizing mutual information is infeasible. Although there is some work on developing surrogate objective functions, most of them delving into the lower bound of mutual information. But these objective functions are not suitable for the minimization, with which the training procedure is easily inclined to divergence. To deal with this problem, we propose to utilize the contrastive log-ratio upper-bound MI estimator \cite{cheng2020club} for achieving disentanglement which estimates the differences of conditional probabilities between positive and negative samples. However, this estimator can not be directly applied as a conditional probability between different components are unavailable such as $p(h_{u,i}|h_{u,j})$, where $i\ne j$. To push this idea forward, we propose to leverage a variational distribution $q_\theta(h_{u,i}|h_{u,j})$ with variational parameter $\theta$ a simple neural network $Q$ to approximate the real conditional one. We define the final object function as:
\begin{equation}
\begin{aligned}
    \mathcal{L}_{mi} &= \sum_i\sum_j \mathbb{E}_{(h_{u,i},h_{u,j})\sim p(h_{u,i},h_{u,j})}[\log q(z_{u,i}|z_{u,j})] \\
     &- \mathbb{E}_{(h_{u,i},h_{u^{\prime},j})\sim p(h_{u,i})p(h_{u,j})}[\log q(z_{u,i}|z_{u^{\prime},j})]
\end{aligned}
\end{equation}

Notice that, $i\ne j$ and $Q$ is trained to minimize the KL-divergence between the true conditional probabilities distribution $p(h_{u,i}|h_{u,j})$ and variational one $q_\theta(h_{u,i}|h_{u,j})$ at the same time.

\begin{equation}
\label{train:q}
    \mathcal{L}_{(h_{u,i},h_{u,j})}=\mathbb{D}_{KL}[p(h_{u,i}|h_{u,j})||q_\theta(h_{u,i}|h_{u,j})]
\end{equation}

Here, we assume the conditional distribution  $p(h_{u,i}|h_{u,j})$ is a Gaussian distribution, and optimize it with final object function alternately.

In summary, explicitly controlling mutual information estimation between components would weak their dependence such that they can capture distinct semantic. \textbf{We refer to \cite{zheng2021adversarial}} and would like to name such disentanglement mechanism as \textit{macro-disentanglement}.

\subsection{Adaptive Scoring}
\label{se:as}
Merit of disentangled knowledge graph representation is its adaption that the importance of each aspect on final prediction can adaptively adjust according to the given scenario (\ie relation in knowledge graph completion). To achieve this goal, in this module (\aka decoder), we first conduct component-level prediction and then leverage an attentive scoring mechanism to guide the fusion of the results from different components.

\textit{Component-level prediction.} Here we take the tail prediction as an example. Once obtained the disentangled head entity and relation representation after the $L$ times neighborhood aggregation, we compute the score for each candidate triplets $(u,r,v)$ in each component. In fact, our method is flexible and can enjoy the merit of recent advances in knowledge graph completion --- we can choose any score function that recent work delving into, such as transE \cite{bordes2013translating}, DistMult \cite{yang2014embedding} and ConvE \cite{dettmers2018convolutional}. Here we take ConvE as an example for description:
\begin{equation}
    \psi_{(u,r,v)}^k=f(vec(f(\overline{h_{u,k}^L};\overline{h_r^L}\star \omega))W)h_{v,k}^L
\end{equation}
Where, $\overline{h_{u,k}^L} \in \mathbb{R}^{d_w \times d_h}, \overline{h_{r}^L} \in \mathbb{R}^{d_w \times d_h}$ denote 2D reshaping of ${h_{v,k}^L} \in \mathbb{R}^{d_w d_h \times 1}, {h_{r}^L} \in \mathbb{R}^{d_w d_h \times 1}$, and $(\star)$ denotes the convolution operation. The whole formula represents the plausibility of the given triplet $(u,r,v)$ in the $k$-th component.

\textit{Relation-aware attentive fusion.} To make our model is adaptive to the given relation, we add the attention scoring $\beta_{(u,r)}^k$  after the decoder module. We assume that, in the relation-aware subspace, the best-matched component representation should be closer to the given relation embedding. In other words, As for query \textit{(Kobe Byrant, member of the sports team, ?)}, our model should pay more attention to the component which aggregates more neighboring information from the topic "career". And the "career" component contains multi triplets such as \textit{(Kobe, sports number, 24), (Kobe, position, shooting guard)} will be "closer" to the relation \textit{member of sports team} against other components. Hence, by sharing the mapping matrix $\theta_r$ with the relation-aware aggregation module, it can facilitate disentangling components thoroughly and distinguish the correct triplet.

\begin{equation}
\begin{aligned}
    \beta_{(u,r)}^k&=softmax((h_{u,k}^L\circ \theta_r)^T \cdot h_r^L) \\
    &=\frac{exp((h_{u,k}^L \circ \theta_r)^T \cdot h_r^L)}{\sum_{k^{\prime}} exp(h_{u,{k^{\prime}}}\circ \theta_r)^T \cdot h_r^L)}
\end{aligned}
\end{equation}

where $h_{u,k}^L$ and $h_r^L$ denotes the final representation of $k$-th component of entity $u$ and relation $r$ respectively.

\begin{equation}
    \psi_{(u,r,v)}^{final} = \sum_k \beta_{(u,r)}^k\psi_{(u,r,v)}^k
\end{equation}

\subsection{Full Objective}
In the training procedure, we leverage a standard cross entropy loss with label smoothing, defined as:
\begin{equation}
\begin{aligned}
    \mathcal{L}=&-\frac{1}{B}\frac{1}{N}\sum_{(u,r)\in batch}\sum_i(t_i\cdot \log(\psi_{(u,r,v_i)}^{final}) \\
    &+ (1-t_i)\cdot \log(1-\psi_{(u,r,v_i)}^{final}))+\lambda\cdot\mathcal{L}_{mi}
\end{aligned}
\end{equation}

Where $B$ is the batch size, $N$ is the number of entities in KG, $t_i$ is the label of the given query (u,r), and $\mathcal{L}_{mi}$ and $\lambda$ are the mutual information regularization loss and its corresponding hyper-parameter. Note that, $Q$ in Formula (\ref{train:q}) is optimized in an alternative way during the training procedure.
\begin{table*}[t]
    \caption{KG completion results for embeddings on FB15k-237 and WN18RR. The best results are in bold and the second best results are underlined.}
    \label{tab:overal_performance}
    \centering
    \begin{tabular}{l|ccccc|ccccc}
        \toprule
        
        \multirow{2}{*}{Model} & \multicolumn{5}{c|}{FB15k-237} & \multicolumn{5}{c}{WN18RR} \\
         & MRR & MR & Hits@1 & Hit@3 & Hit@10 & MRR & MR & Hits@1 & Hit@3 & Hit@10  \\
        \midrule
        TransE \cite{bordes2013translating} & 0.294 & 357 & - & -  & 0.465 & 0.226 & 3384 & - & - & 0.501 \\
        Distmult \cite{yang2014embedding} & 0.241 & 254 & 0.155 & 0.263  & 0.419 & 0.43 & 5110 & 0.39 & 0.44 & 0.49 \\
        ConvE \cite{dettmers2018convolutional} & 0.325 & 244 & 0.237 & 0.356 & 0.501 & 0.43 & 4187 & 0.40 & 0.44 & 0.52 \\
        RotatE \cite{sun2019rotate} & 0.338 & \underline{177} & 0.241 & 0.375  & 0.533 & 0.476 & 3340 & 0.428 & 0.492 & 0.571 \\
        SACN \cite{shang2019end} & 0.35 & - & 0.261 & 0.39  & 0.54 & 0.47 & - & 0.43 & 0.48 & \underline{0.54} \\
        InteractE \cite{vashishth2020interacte} & 0.354 & \textbf{172} & 0.263 & - & 0.535 & 0.463 & 5202 & 0.43 & - & 0.528 \\
        MuRE \cite{balazevic2019multi} & 0.336 & - & 0.245 & 0.370 & 0.521 & 0.465 & - & 0.436 & 0.487 & 0.554  \\
        COMPGCN \cite{vashishth2019composition} & 0.355 & 197 & 0.264 & 0.39 & 0.535 & \underline{0.479} & 3533 & \textbf{0.443} & \underline{0.494} & 0.546 \\
        AcrE \cite{ren-etal-2020-knowledge} & \underline{0.358} & - & \underline{0.266} & \underline{0.393} & \underline{0.545} & 0.459 & - & 0.422 &0.473 & 0.532 \\
        ReInceptionE \cite{xie-etal-2020-reinceptione} & 0.349 & 173 & - & - & 0.528 & 0.483 & \underline{1894} & - & - & 0.582  \\
        
        \hline
        DisenKGAT & \textbf{0.368} & 179 & \textbf{0.275} & \textbf{0.407} & \textbf{0.553} & \textbf{0.486} & \textbf{1504} & \underline{0.441} & \textbf{0.502} & \textbf{0.578}  \\
        
        \bottomrule
    \end{tabular}

\end{table*}
\begin{table}[t]
    \small
    \caption{Dataset statistics}
    \label{tab:datasets}
    \centering
    \begin{tabular}{cccccc}
        \toprule
        Data sets & $|\varepsilon|$ & $|\mathcal{R}|$ & \multicolumn{3}{c}{Triplets} \\
        \cline{4-6}
        &&& Train & Valid & Test \\
        \hline
        FB15k-237 & 14,541 & 237 & 272,114 & 17,535 & 20,466 \\
        WN18RR & 40,943 & 11 & 86,835 & 3,034 & 3,134 \\
        \bottomrule
    \end{tabular}

\end{table}



\section{Experiments}
In this section, we present experiments to demonstrate the effectiveness of DisenKGAT. We design numerous experiments aim to answer the following research questions:
\begin{itemize}[leftmargin=*]
    \item \textbf{RQ1:} How does DisenKGAT perform compared to existing aproaches, \textit{w.r.t.} distance-based and semantic matching models?
    \item \textbf{RQ2:} How do the critical components~(e.g., relation-aware aggregation) contirbute to DisenKGAT and how do different hyper-parameters~(e.g., factor number) affect DisenKGAT?
    \item \textbf{RQ3:} Does DisenKGAT work roubstly with other decoder modules?
    \item \textbf{RQ4:} Can DisenKGAT give explanations of the benefits brought by the disentangled factors?
\end{itemize}

\subsection{Experimental Settings}
\subsubsection{\textbf{Data Sets}}
To verify the performance of DisenKGAT, we select two most commonly used public datasets: FB15k-237 \cite{toutanova2015observed} and WN18RR\cite{dettmers2018convolutional}. The summary statistics of the above dataset is listed in Table.
FB15k-237 is an improved version of the FB15k \cite{bordes2013translating} dataset where all trivial inverse relations are moved. It contains large-scale general knowledge facts such as awards, places, and actors.
WN18RR is a subset of WN18\cite{bordes2013translating} that stores structured knowledge of English lexicons including hypernym, hyponym, and synonym, etc. It also deletes inverse relations similar to FB15k-237.

\subsubsection{\textbf{Baselines}}
We compare our DisenKGAT with several state-of-the-art methods, including distanced-based models (TransE \cite{bordes2013translating}, RotatE \cite{sun2019rotate}), semantic matching models (Distmult \cite{yang2014embedding}, RESCAL), and neural network based models ( ConvE \cite{dettmers2018convolutional}, InteractE \cite{vashishth2020interacte}, SACN \cite{shang2019end}, ArcE \cite{ren-etal-2020-knowledge}, ReinceptionE \cite{xie-etal-2020-reinceptione}, COMPGCN \cite{vashishth2019composition}).

\begin{itemize}[leftmargin=*]
    \item \textbf{TransE} \cite{bordes2013translating} is the first translation based work in knowledge graph embedding by assuming that the superposition of head and relation embedding $h+r$ should be close to the tail embedding of $t$.
    \item \textbf{RESCAL} \cite{nickel2011three} is the three-way rank-$r$ factorization methods which treats each relation as a full rank matrix.
    \item \textbf{DistMult} \cite{yang2014embedding} puts forward a simplified bilinear formulation which deals with each relation as a diagonal matrix.
    \item \textbf{InteractE} \cite{vashishth2020interacte} is an extensive version of ConvE. It increases the rich interactions between entities and relations by the usage of feature permutation and circular convolution.
    \item \textbf{RotatE} \cite{sun2019rotate} is built upon TransE, which regards relation as a kind of rotation from head entiy to tail entity.
    \item \textbf{ConvE} \cite{dettmers2018convolutional} is a representative neural network based models, which leverage 2D convolution and multiple layers of nonlinear transformation to model the complex interaction between entities and relations.
    \item \textbf{SACN} \cite{shang2019end} proposes an end-to-end structure aware convolutional Network that takes the benefit of GCN and ConvE together.
    \item \textbf{MuRE} \cite{balazevic2019multi} is translational distance hyperbolic embedding with a diagonal relational matrix.
    \item \textbf{ArcE} \cite{ren-etal-2020-knowledge} utilizes a simple but effective atrous convolution on knowledge graph embedding.
    \item \textbf{ReinceptionE} \cite{xie-etal-2020-reinceptione} proposes an inception network with joint local-global structural information.
    \item \textbf{COMPGCN} \cite{vashishth2019composition} is a general framework which incorporates graph neural network with knowledge graph embedding.
\end{itemize}

\begin{table*}
    \caption{Results on link prediction by relation category on FB15k-237 dataset. Following the \cite{wang2014knowledge}, the relations are divided into four categories: one-to-one (1-1), one-to-many (1-N), many- to-one (N-1), and many-to-many (N-N).}
    \label{tab:1-n}
    \centering
    \begin{tabular}{ccccccccccccc}
        \toprule
        \multicolumn{2}{c}{\multirow{2}{*}{}}& \multicolumn{2}{c}{RotatE} &&  \multicolumn{2}{c}{WGCN} && 
        \multicolumn{2}{c}{COMPGCN} && \multicolumn{2}{c}{DisenKGAT}\\
        \cline{3-4} \cline{6-7} \cline{9-10} \cline{12-13}
        \multicolumn{2}{c}{} & MRR  & H@10  &&MRR  & H@10 && MRR  & H@10 && MRR  & H@10 \\
        \hline
        \multirow{4}{*}{Head Pred} & 1-1 & 0.498 & 0.593 && 0.422 & 0.547 && 0.457  & 0.604 && 
        \textbf{0.501} &  \textbf{0.625} \\
        & 1-N & 0.092  & 0.174 && 0.093 & 0.187 && 0.112 & 0.190 & &
        \textbf{0.128}  & \textbf{0.248} \\
        & N-1 & 0.471  & 0.674 && 0.454 & 0.647 && 0.471 & 0.656 && 
        \textbf{0.486}  & \textbf{0.659} \\
        & N-N & 0.261  & 0.476 && 0.261 & 0.459 && 0.275 & 0.474 & &
        \textbf{0.291} & \textbf{0.496} \\
        \hline
        \multirow{4}{*}{Tail Pred} & 1-1 & 0.484 & 0.578 && 0.406 & 0.531 && 0.453 & 0.589 &&
        \textbf{0.499}  & \textbf{0.641} \\
        & 1-N & 0.749 & 0.674  && 0.771 & 0.875 && 0.779  & 0.885 &&
        \textbf{0.789}  & \textbf{0.889} \\
        & N-1 & 0.074  & 0.138 && 0.068 & 0.139 && 0.076  & 0.151 && 
        \textbf{0.086}  & \textbf{0.180} \\
        & N-N & 0.364  & 0.608 && 0.385 & 0.607 && 0.395  & 0.616 && 
        \textbf{0.402}  & \textbf{0.629} \\
        \bottomrule
    \end{tabular}
\end{table*}

\subsubsection{\textbf{Evaluation protocol}}
We perform the link prediction task following the settings in \cite{bordes2013translating}, given the test triplet to be predicted, we rank it against all candidate triplets which filter out all the valid triplets appearing in the training, validation, and test datasets. And inverse relation is also added to predict the head entity \cite{pmlr-v80-lacroix18a} during the training procedure. Meanwhile, the performance of our model is reported on the standard link prediction metrics: Mean Reciprocal Rank (MRR), Mean Rank (MR), and Hits@1, Hits@3, Hits@10.

\subsubsection{\textbf{Paramter Settings}}
We implement our DisenKGAT model in PyTorch and have released our implementation (codes, parameter setting, and training log) to enhance the reproducibility. In our method, each component embedding size is fixed to 200 for a fair comparison. And we fix the optimizer as Adam. A grid search is conducted to confirm the optimal parameter setting for each model. To be more specific, the number of layer is tuned amongst \{1, 2, 3\}, dropout is tuned in \{0.0, 0.1, ...0.5\} to prevent over-fitting, batch size is varying in \{128, 256, 512, 1024, 2048\}, and the coefficients of Mutual information regularization are search in \{0.1, 0.01, 0.001, 0.0001\}. We initialize the model parameter with Xavier for a fair comparison with GNN-based models, especially for COMPGCN.

\subsection{Performance Comparsion (RQ1)}
We begin to summarize the results by analyzing the performance of all models reported in table and have the following observations:
\begin{itemize}[leftmargin=*]
\item From the results in Table \ref{tab:overal_performance}, we can see our proposed DisenKGAT obtains competitive results compared with state-of-the-art models. In particular, DisenKGAT achieve a considerable improvement on the FB15k-237 dataset w.r.t MRR, Hits@1, and Hits@3, it is also superior to other models using MR on the WN18RR dataset. We attribute these results to the disentanglement module on a complex knowledge graph. Since the FB15k-237 dataset includes 237 relation types, it is suitable for our motivation to deal with multi-relation entanglement issues. Meanwhile, WN18RR only contains 11 relation types, and each entity embodies an extremely single meaning. For example, the entity "pass on" has three diverse meanings (transmit, transfer possession of, cause be distributed), and each meaning belongs to an individual entity. Hence, our latter experiments will pay more attention to FB15k-237.
    \item Focus on the complex multi-relations scenario especially for the issue of 1-N, N-1, and N-N relations (shown in Table \ref{tab:1-n}), We present the experimental results on different relation types following the \cite{wang2014knowledge}. We choose the FB15k-237 as our object owing to its abundant multi-relations and denser graph structure and compare our model with ROTATE \cite{sun2019rotate}, W-GCN \cite{shang2019end}, COMPGCN \cite{vashishth2019composition}. We find that DisenKGAT outperforms baselines in all relation types. More specifically, GNN-based models (W-GCN, COMPGCN) are superior to RotatE on complex relation types (1-N, N-1, N-N) which indicates that full utilization of graph structure facilitates handling complex relations. Meanwhile, 
    we observe that RoatatE outperforms W-GCN and COMPGCN on simple relation (1-1), which can be attributed to its strong capability in capturing various relation structures, including symmetry/antisymmetry, composition, and inversion. Just as our motivation, we tend to capture rich latent factors in complex scenarios and generate adaptive embeddings in terms of the given scenario. Therefore, our model will aggregate those neighbors which are in similar "topics" potentially, thus outperforms other models by a large margin in both simple and complex relations.
\end{itemize}

\begin{table}[t]
    \small
    \caption{Result of ablation study}
    \label{tab:ablation}
    \centering
    \begin{tabular}{cccccc}
        \toprule
        model & MRR & MR & Hits@1 & Hits@3 & Hits@10 \\
        \hline 
         w/o micro & 0.355 & 197& 0.265 & 0.392 & 0.534 \\
        \hline
        w/o macro & 0.356 & 303 & 0.263 & 0.392 & 0.542 \\
        w/o HSIC & 0.352 & 259 & 0.263 & 0.387 & 0.527 \\
        \hline
        \hline
        DisenKGAT & 0.368 & 179 & 0.275 & 0.407 & 0.553\\
        \bottomrule
    \end{tabular}

\end{table}
\subsection{Ablation Study (RQ2)}

As the disentangled propagation mode consistently outperforms all kinds of baseline along with various score functions, we investigate the impact of each module of the model to analyze it deeply and comprehensively. To be more specific, we conduct ablation studies on the presence of aggregation mechanism, independence modeling, attention scoring, and several hyper-parameters.
\begin{table*}[ht]
    \caption{Performance on link prediction task evaluated on FB15k-237 dataset. Similar to COMPGCN, X + M (Y) denotes that method M is used for obtaining entity (and relation) embeddings with X as the scoring function. Y denote the composition operator $\phi$. And detailed analysis refer to section }
    \label{tab:replace_score}
    \centering
    \begin{tabular}{lccccccccccc}
        \toprule
        \textbf{Scoring Function}(=X)$\to$ & \multicolumn{3}{c}{\textbf{TransE}} & &
        \multicolumn{3}{c}{\textbf{DistMult}} && \multicolumn{3}{c}{\textbf{ConvE}} \\
         \cline{2-4}  \cline{6-8}  \cline{10-12}  \\
        \textbf{Methods}$\downarrow$ & MRR & MR & H@10 && MRR & MR & H@10 && MRR & MR & H@10 \\
        \hline
        X & 0.294 & 357 & 0.465 && 0.241 & 354 & 0.419 && 0.325 & 244 & 0.501 \\
        X+D-GCN & 0.299 & 351 & 0.469 && 0.321 & 225 & 0.497 && 0.344 & 200 & 0.524 \\
        X+W-GCN & 0.264 & 1520 & 0.444 && 0.324 & 229 & 0.504 && 0.244 & 201 & 0.525 \\
        \hline
        X+COMPGCN(sub) & 0.335 & 194 & 0.514 && 0.336 & 231 & 0.513 && 0.352 & 199 & 0.530 \\
        X+COMPGCN(Mult) & 0.337 & 233 & 0.515 && 0.338 & 200 & 0.518 && 0.353 & 216 & 0.532 \\
        X+COMPGCN(Corr) & 0.336 & 214 & 0.518 && 0.335 & 227 & 0.514 && 0.355 & 197 & 0.535 \\
        \hline
        X+DisenKGAT(sub) & 0.334 & \underline{183} & 0.51 && 0.346  & \underline{196} &0.531  && 0.358 & 181 & 0.543 \\
        X+DisenKGAT(Mult) & \underline{0.342} & \textbf{170} & \underline{0.524} &&  \underline{0.353} & \textbf{184} & \underline{0.536} && \underline{0.364} & \textbf{171} & \underline{0.550} \\
        X+DisenKGAT(Corr) & 0.338 & 203 & 0.520 && 0.341  & 200 & 0.528 && 0.359 & 189 & 0.541 \\
        X+DisenKGAT(Cross) & \textbf{0.343} & 187 & \textbf{0.526} && \textbf{0.354}  & 204 & \textbf{0.540} && \textbf{0.368} & \underline{179} & \textbf{0.553} \\
        \bottomrule
    \end{tabular}
\end{table*}
\subsubsection{Impact of micro-disentanglement}

To achieve the micro-disen\\tanglement, we utilize a relation-aware aggregation module which facilitates the distinction of different components of each entity. So is it crucial to consider neighboring information and utilize such propagation layers? Toward this end, multi variants  are constructed by removing the whole micro aggregation module termed as $\rm{DisenKGAT}_{w/o\,micro}$.The detailed results are listed in Table \ref{tab:ablation}. 
In contrast to the complete model, $\rm{DisenKGAT}_{w/o\,micro}$ obtains dramatically worse results, which can be attributed to the following reasons. (1) Owing to the operator of relation-aware mapping, each component is projected into different latent subspace and contains individual semantic information.(2) With the assistance of relation-aware aggregation, our model explores the deeply fine-grained level of "topics" and collect specific neighbors in different scenario. However, lacking the micro-disentanglement, our model degrade into COMPGCN, which leaving entanglement unexplored.


\subsubsection{Impact of macro-disentanglement}

To achieve the macro-dise-\\ntanglement, we leverage mutual information minimization to ensure each component be separate. Intuitively, without the separability regularization, it is prone to cause the aggregated information to be unrepresentative in terms of the given scenario. In particular, we conduct experiments to verify our intuition. (1) On the one hand, removing the mutual information regularization, denoted as ($\rm{DisenKGAT}_{w/o\,macro}$), helps to validate the importance of macro-separability. (2) On the other hand, we replace the MI regularization with Hilbert-Schmidt Independence Criterion (HSIC), denoted as $\rm{DisenKGAT}_{w/o \,HSIC}$, which is also common metrics measuring two-variable correlation.

Obviously, the non-regularization mode is inferior to the complete model. We suspect that mutual information helps to generate diverse component representation to some extent and avoids producing the representation of repeated semantics.  Furthermore, the choice of macro-disentanglement regularization is also critical in the training procedure. HSIC is also a common measurement of correlation which was utilized in some tiny homogeneous graphs \cite{liu2020independence}. However, the empirical result shows that it is not suitable for more complex heterogeneous graphs such as our knowledge graph.

\subsubsection{Impact of component number}

\begin{figure}[tp] 

    \centering
    \subfloat[FB1515k-237]{%
        \includegraphics[width=0.236\textwidth]{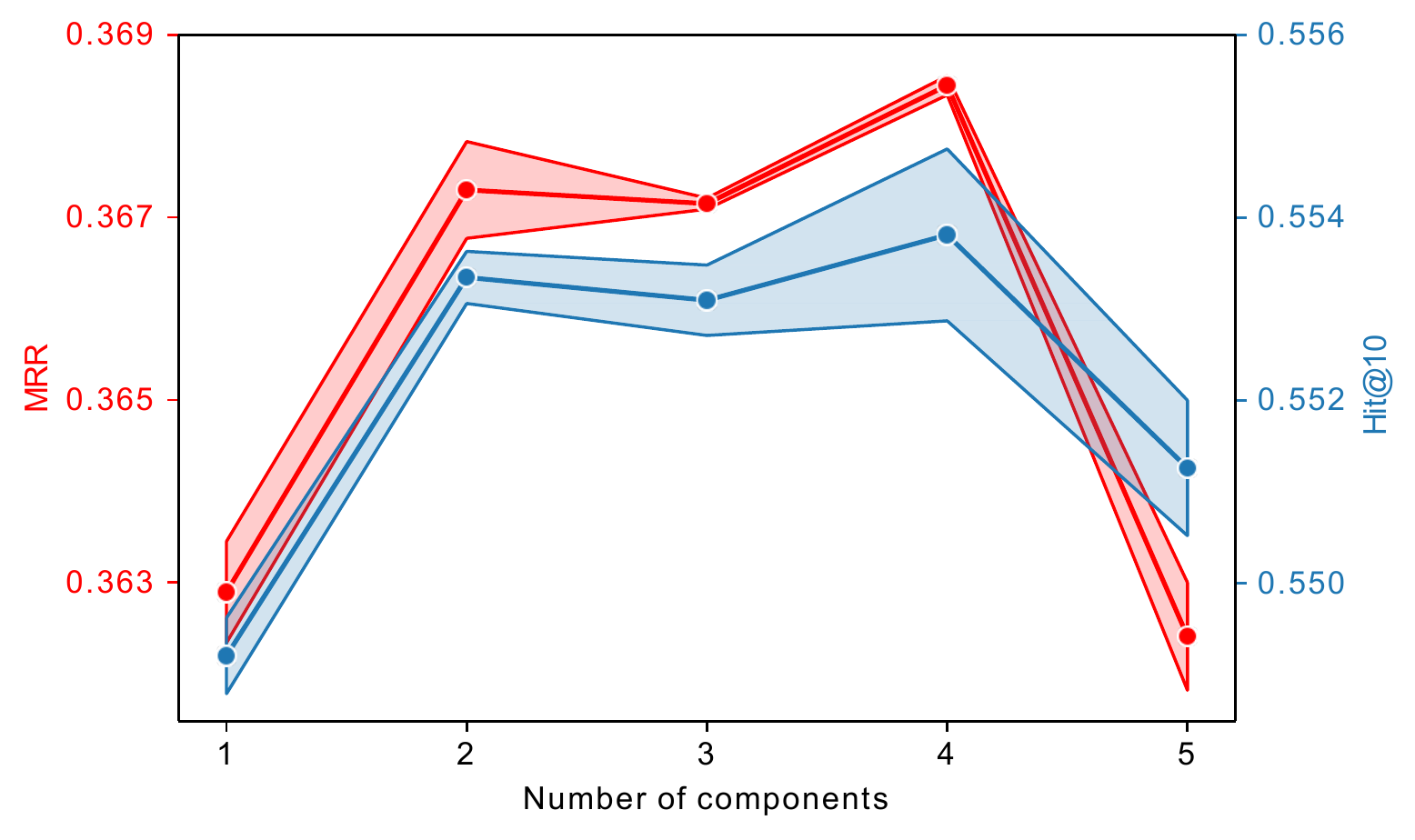}%
        \label{fig:a}%
        }%
    \hfill%
    \subfloat[WN18RR]{%
        \includegraphics[width=0.236\textwidth]{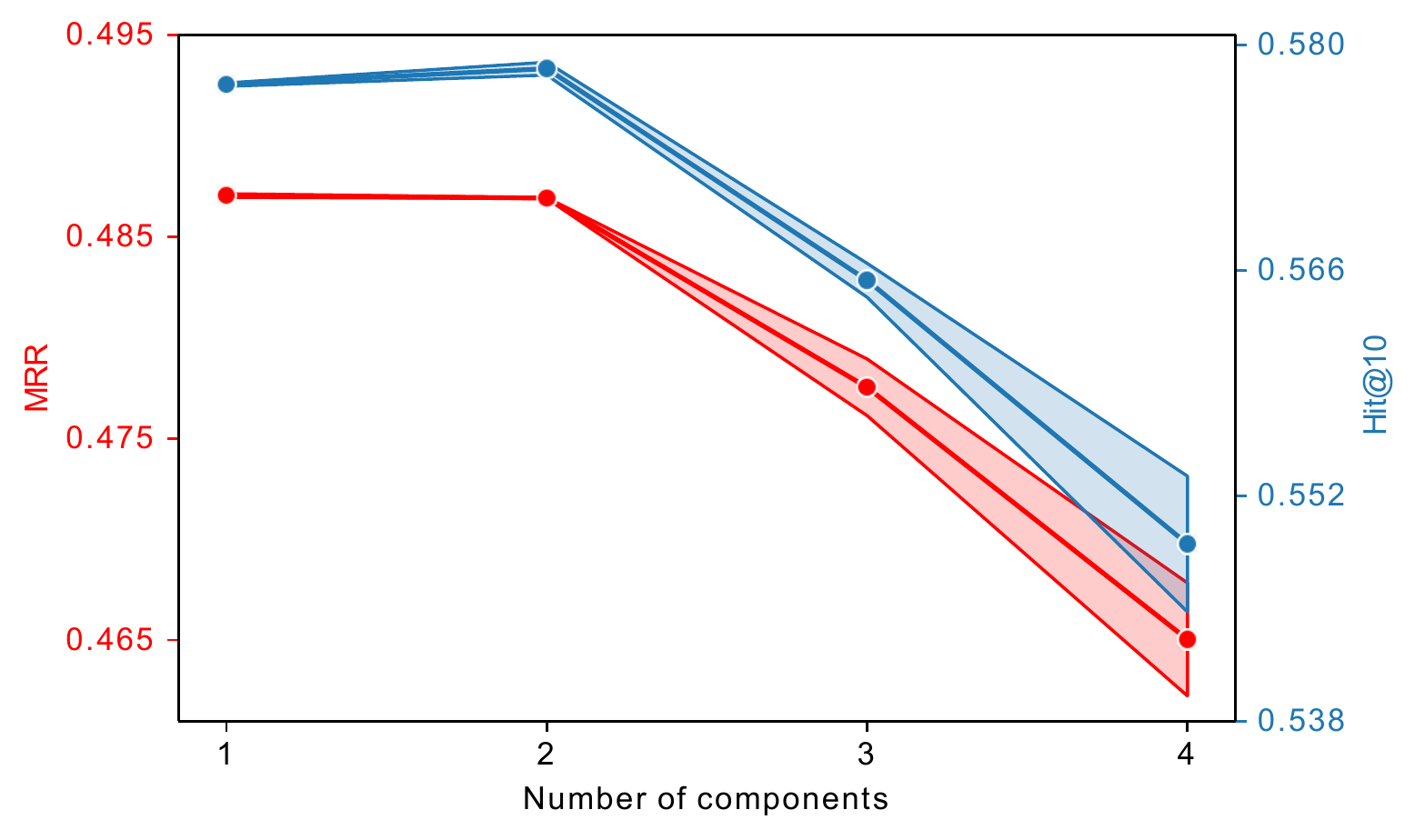}%
        \label{fig:b}%
        }%
    \caption{Impact of component number ($K$).}
    \label{fig:k}
\end{figure}

In this section, we consider varying the number of disentangled factors. The suitable disentangled factors are related to the facts in the real datasets.  Here we search the component numbers in the range of \{1, 2, 3, 4, 5\}. Figure \ref{fig:k} summarizes the experimental results \textit{w.r.t.} the FB15k-237 and WN18RR data set. We observe that:

When the choice of the number of disentangled components is 1 in FB15k-237, our model intuitively degrades into a normal GNN-based model which coupling with attention aggregation and relation-aware mapping. Although it outperforms the COMPGCN  attributed to the flexible aggregation mechanism, it still leads to suboptimal results. We suspect that lacking exploiting the latent factors will be difficult to deal with complex semantic meanings and degenerate to static representation learning. 

In FB15k-237, increasing the number of disentangled components can substantially provide a considerable improvement in terms of all measures. The best choice is around 4, which is related to the "topics number" of most entities. When $k>4$, the performance dramatically degrades since it makes some topics too fine-grained to carry crucial information. The difference is that in the WN18RR dataset, because of its simple and fine-grained meaning, the best selection of K is about 2, and with the increase of K value, the 
performance collapses significantly.

\subsection{Robustness (RQ3)}
\label{sec:rob}
As mentioned in contribution, our model has strong robustness in knowledge graph embedding since it could work along with multi-categories score function and has a considerable improvement. Borrowed from COMPGCN, we evaluate the effect of using different representative score function (shown in Table \ref{tab:replace_score}) : TransE (translational), DistMult (semantic matching), and ConvE (neural-based model). Inspired from \cite{bordes2013translating, schlichtkrull2018modeling, zhang2019interaction}, we define four kinds of operators:
\begin{itemize}[leftmargin=*]
    \item \textbf{Subtraction(Sub)} : $\phi(e_s,e_r,\theta)=\, (\theta_r \cdot e_s)-e_r$
    \item \textbf{Multiplication(Mult)} : $\phi(e_s,e_r,\theta)=\, (\theta_r\cdot e_s)\circ r_r$
    \item \textbf{circular correlation(Corr)} : $\phi(e_s,e_r,\theta)=\, (\theta_re_s) \star e_r $
    \item \textbf{Crossover Interaction(Cross)} : \\ $\phi(e_s,e_r,\theta)=\, \theta_r e_s+\theta_r(e_s\circ e_r)$
\end{itemize}
The empirical results are shown in Table \ref{tab:replace_score}. We observe that:

\begin{itemize}[leftmargin=*]
    \item By the utilization of the Graph Convolution Network, it can further explore the graph structural information. Hence it could provides a substantial improvement. And we suspect that owing to the lacking consideration of incorporating relation embedding into the propagation layer at the aggregation stage, W-GCN and D-GCN do not perform well with the various score functions. On the contrary, COMPGCN and our proposed DisenKGAT outperform across the three score functions in terms of all measures.
    \item On average, our DisenKGAT obtains around 2\%, 5\%, 3\% relative increase in MRR with TransE, DistMult, and ConvE respectively compare to the COMPGCN. To be more specific, as the design of relation-aware aggregation and mutual information regularization empowers the model with rich structural information and disentangled latent factors, we can surpass the baseline models by a large margin in all metrics. However, unlike other combinations, our model with TransE score function in "sub" composition operator leads to slight improvement. We attribute these results to the shallowness of the intrinsic structure used in TransE, which is difficult to capture the deeply complex interactions between entities and relations.
\end{itemize}

\subsection{Explainability of DisenKGAT (RQ4)}
\label{sec:vis}
\begin{figure}[t] 
    \centering 
    \includegraphics[scale=0.23]{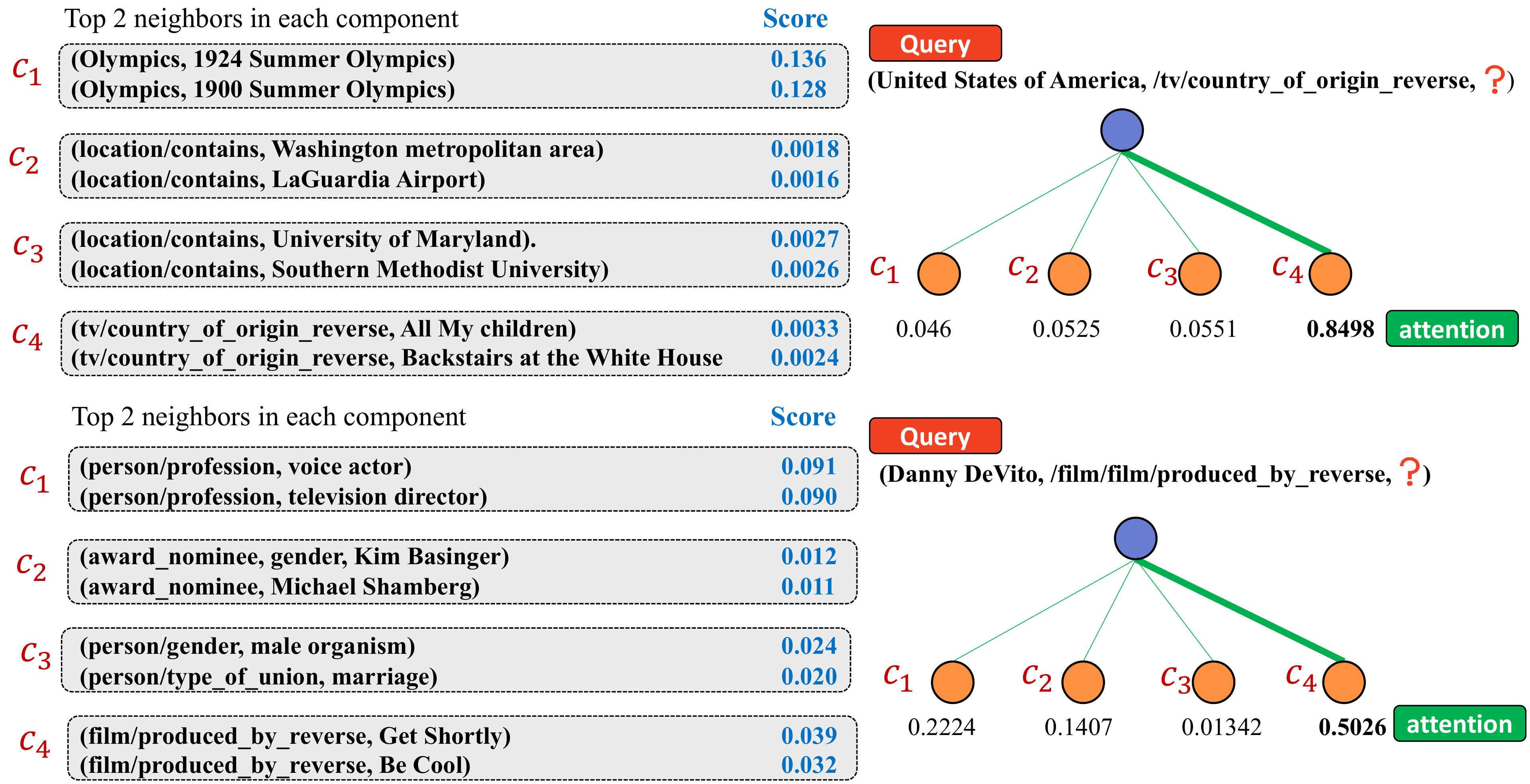} 
    \caption{ Explanations of disentangled performance. Best viewed in color. } 
    \label{fig::explan} 
\end{figure}

In this section, we exploit the disentangled component of each entity in the test set to investigate how the disentangled factors facilitate the embedding. We provide two examples from FB15k-237 to show an intuitive impression of our disentangled performance. The detailed results are displayed in Figure \ref{fig::explan}, we then have the following observations:
\begin{itemize}[leftmargin=*]
    \item In the module of relation-aware aggregation, we leverage a similarity score to measure the correlation between neighbors in each component. We assume that, the similarity between the entity $u$ and $v$ in the $k$-th component subspace represents the importance of factor $k$ in this scenario. In other words, factor $k$ may be associated with a "topic" or a "cluster" which influences the reason why $u$ and $v$ are linked. For example in the upper part of Figure \ref{fig::explan}, the top two neighbors of entity "United States of American" in component $c_1$ are \textit{(Olympics, 1924 Summer Olympics)}\footnote{Note that the complete triplet is \textit{(United States of American, Olympics, 1924 Summer Olympics)}. We discuss the neighbors here in the format of (relation, tail) as above.} and \textit{(Olympics, 1900 Summer Olympics)}, while \textit{(location/contains, Washington metropolitan area)} and \textit{(location/contains, LaGuardia Airport)} are ranked top two in component $c_2$. It is clear to figure out that, each component of the entity "United States of American" plays a diverse role and constructs distinguishable clusters potentially.
    \item In the module on attention scoring, we utilize a similarity score $\beta$ to match the given query to the semantic representation in each component. As shown in the right part of Figure \ref{fig::explan}, given the query of \textit{(United States of the America, /tv/country\_of\_origin\_reve-\\rse,?)}, the component $c_4$ rank first in the whole semantic space. In terms of the analysis in the last point, we could guess component $c_1, c_2, c_3, c_4$ are related to the topic "Sports", "place", "City" and "film and TV" respectively. Obviously, $c_4$ is closer to the given query, and its weight reflects the important influence in the task of link prediction. 
    \item As the whole computation is element-wise, link prediction is under the scoring of the triplets in each component independently. So the topic or cluster in each component of various entities should be shared all the time. We find that given the query \textit{(Danny DeVito, /film/produced\_by\_inverse)}, the most crucial component is also $c_4$.  Similar to the analysis in point 2 above, the component $c_4$ about the entity "Danny DeVito" is about "film" which is similar to the entity "The United States of the America". This empirically shows that the entity "Danny DeVito" may be linked with "United States of the America" in some scenarios such as "nationality". Then the model induces a similar topic $c_4$ in both entities which intuitively helps to deal with the embedding in the complex knowledge graph.

\end{itemize}
\section{CONClUSION AND FUTURE WORK}
In this work, we focus on the disentangled representation behind entities in the knowledge graph for link prediction. We propose a novel Disentangled Knowledge attention network, DisenKGAT, which exploits the disentangled factors from two views: (1) leveraging a novel relation-aware aggregation mechanism to achieve the micro-disentanglement while ensuring the structural information; and (2) taking the macro-separability into consideration, which utilizes mutual information minimization as a regularization. Furthermore, we provide various experiments to verify the effectiveness and explainability of our model.

For future work, we will explore a more general disentangled framework that could adapt to more complex scenarios. For example, how to disentangle knowledge graph in a fine-grained level, where the number of components per entity can be different and flexible. Besides, we also have a plan to explore the potential of mutual information in more various scenarios rather than just as a regularization in our work.

    
    

\begin{acks}
This work is supported by the National Key Research and Development Program of China (2020AAA0106000), USTC Research Funds of the Double First-Class Initiative (WK2100000019), and the Meituan.
\end{acks}

\bibliographystyle{ACM-Reference-Format}
\bibliography{Disen}

\end{document}